\documentclass[letterpaper]{article}
\usepackage{ijcai13}

\usepackage{times}
\usepackage[named]{algo}
\usepackage{wrapfig}
\usepackage{epsfig}
\usepackage{amssymb}
\usepackage{amsmath}
\usepackage{amsfonts}
\usepackage{ifthen}

\usepackage{graphicx}
\usepackage{caption}
\usepackage{subcaption}

\newcommand{\argmax}{ \operatorname*{arg\,max}}
\newtheorem{theorem}{Theorem}

\newcommand{\squishlist}{
 \begin{list}{$\bullet$}
  { \setlength{\itemsep}{5pt}
     \setlength{\parsep}{0pt}
     \setlength{\topsep}{0pt}
     \setlength{\partopsep}{0pt}
     \setlength{\leftmargin}{0.7em}
     \setlength{\labelwidth}{0.5em}
     \setlength{\labelsep}{0.2em} } }

\newcommand{\squishlisttwo}{
 \begin{list}{$\bullet$}
  { \setlength{\itemsep}{2pt}
     \setlength{\parsep}{0pt}
    \setlength{\topsep}{2pt}
    \setlength{\partopsep}{0pt}
    \setlength{\leftmargin}{1em}
    \setlength{\labelwidth}{1.5em}
    \setlength{\labelsep}{0.5em} } }

\newcommand{\squishend}{
  \end{list}  }

\title{A General Framework for Interacting Bayes-Optimally with\\Self-Interested Agents using Arbitrary Parametric Model and Model Prior}
\author{
Trong Nghia Hoang \and Kian Hsiang Low\\
Department of Computer Science, National University of Singapore\\
Republic of Singapore\\
\{nghiaht, lowkh\}@comp.nus.edu.sg}

\begin{document}

\newcounter{sol} 
\setcounter{sol}{1} 

\maketitle

\begin{abstract}
Recent advances in Bayesian reinforcement learning (BRL) have shown that Bayes-optimality is theoretically achievable by modeling the environment's latent dynamics using Flat-Dirichlet-Multinomial (FDM) prior. In self-interested multi-agent environments, the transition dynamics are mainly controlled by the other agent's stochastic behavior for which FDM's independence and modeling assumptions do not hold. As a result, FDM does not allow the other agent's behavior to be generalized across different states nor specified using prior domain knowledge. To overcome these practical limitations of FDM, we propose a generalization of BRL to integrate the general class of parametric models and model priors, thus allowing practitioners' domain knowledge to be exploited to produce a fine-grained and compact representation of the other agent's behavior. Empirical evaluation shows that our approach outperforms existing multi-agent reinforcement learning algorithms.
\end{abstract}

\section{Introduction}
\label{sect:intro}
In reinforcement learning (RL), an agent faces a dilemma between acting optimally with respect to the current, possibly incomplete knowledge of the environment (i.e., exploitation) vs. acting sub-optimally to gain more information about it (i.e., exploration). Model-based Bayesian reinforcement learning (BRL) circumvents such a dilemma by considering the notion of Bayes-optimality \cite{Duff2003}: A Bayes-optimal policy selects actions that maximize the agent's expected utility with respect to all possible sequences of future beliefs (starting from the initial belief) over candidate models of the environment. Unfortunately, due to the large belief space, the Bayes-optimal policy can only be approximately derived under a simple choice of models and model priors. For example, the Flat-Dirichlet-Multinomial (FDM) prior \cite{Poupart2006} assumes the next-state distributions for each action-state pair to be modeled as independent multinomial distributions with separate Dirichlet priors. Despite its common use to analyze and benchmark algorithms, FDM can perform poorly in practice as it often fails to exploit the structured information of a problem \cite{Asmuth2011,Mauricio2012}.

To elaborate, a critical limitation of FDM lies in its independence assumption driven by computational convenience rather than scientific insight. We can identify practical examples in the context of self-interested multi-agent RL (MARL) where the uncertainty in the transition model is mainly caused by the stochasticity in the other agent's behavior (in different states) for which the independence assumption does not hold (e.g., motion behavior of pedestrians [Natarajan \emph{et al.}, 2012a; 2012b]). Consider, for example, an application of BRL in the problem of placing static sensors to monitor an environmental phenomenon: It involves actively selecting sensor locations (i.e., states) for measurement such that the sum of predictive variances at the unobserved locations is minimized. Here, the phenomenon is the ``other agent" and the measurements are its actions. An important characterization of the phenomenon is that of the spatial correlation of measurements between neighboring locations/states [Low \emph{et al.}, 2007; 2008; 2009; 2011; 2012; Chen \emph{et al.}, 2012; Cao \emph{et al.}, 2013], which makes FDM-based BRL extremely ill-suited for this problem due to its independence assumption.

Secondly, despite its computational convenience, FDM does not permit generalization across states \cite{Asmuth2011}, thus severely limiting its applicability in practical problems with a large state space where past observations only come from a very limited set of states. Interestingly, in such problems, it is often possible to obtain prior domain knowledge providing a more ``parsimonious" structure of the other agent's behavior, which can potentially resolve the issue of generalization. For example, consider using BRL to derive a Bayes-optimal policy for an autonomous car to navigate successfully among human-driven vehicles [Hoang and Low, 2012; 2013b] whose behaviors in different situations (i.e., states) are governed by a small, consistent set of latent parameters, as demonstrated in the empirical study of \citeauthor{Gipps1981}~\shortcite{Gipps1981}. By estimating/learning these parameters, it is then possible to generalize their behaviors across different states. This, however, contradicts the independence assumption of FDM; in practice, ignoring this results in an inferior performance, as shown in Section~\ref{sect:exp}. Note that, by using parameter tying [Poupart et al., 2006], FDM can be modified to make the other agent's behavior identical in different states. But, this simple generalization is too restrictive for real-world problems like the examples above where the other agent's behavior in different states is not necessarily identical but related via a common set of latent ``non-Dirichlet'' parameters.

Consequently, there is still a huge gap in putting BRL into practice for interacting with self-interested agents of unknown behaviors. To the best of our knowledge, this is first investigated by \citeauthor{Chalkiadakis2003}~\shortcite{Chalkiadakis2003} who offer a myopic solution in the belief space instead of solving for a Bayes-optimal policy that is non-myopic. Their proposed BPVI method essentially selects actions that jointly maximize a heuristic aggregation of myopic value of perfect information \cite{Dearden1998} and an average estimation of expected utility obtained from solving the exact MDPs with respect to samples drawn from the posterior belief of the other agent's behavior. Moreover, BPVI is restricted to work only with Dirichlet priors and multinomial likelihoods (i.e., FDM), which are subject to the above disadvantages in modeling the other agent's behavior. Also, BPVI is demonstrated empirically in the simplest of settings with only a few states.

Furthermore, in light of the above examples, the other agent's behavior often needs to be modeled differently depending on the specific application. Grounding in the context of the BRL framework, either the domain expert struggles to best fit his prior knowledge to the supported set of models and model priors or the agent developer has to re-design the framework to incorporate a new modeling scheme. Arguably, there is no free lunch when it comes to modeling the other agent's behavior across various applications. To cope with this difficulty, the BRL framework should ideally allow a domain expert to freely incorporate his choice of design in modeling the other agent's behavior.

Motivated by the above practical considerations, this paper presents a novel generalization of BRL, which we call \emph{Interactive BRL} (I-BRL) (Section~\ref{sect:ibrl}), to integrate any parametric model and model prior of the other agent's behavior (Section~\ref{sect:model}) specified by domain experts, consequently yielding two advantages: The other agent's behavior can be represented (a) in a fine-grained manner based on the practitioners' prior domain knowledge, and (b) compactly to be generalized across different states, thus overcoming the limitations of FDM. We show how the non-myopic Bayes-optimal policy can be derived analytically by solving I-BRL exactly (Section~\ref{sect:exact}) and propose an approximation algorithm to compute it efficiently in polynomial time (Section~\ref{sect:approx}). Empirically, we evaluate the performance of I-BRL against that of BPVI \cite{Chalkiadakis2003} using an interesting traffic problem modeled after a real-world situation (Section~\ref{sect:exp}).

\section{Modeling the Other Agent}
\label{sect:model}
In our proposed Bayesian modeling paradigm, the opponent's\footnote{For convenience, we will use the terms the ``other agent'' and ``opponent'' interchangeably from now on.} behavior is modeled as a set of probabilities $p_{sh}^{v}(\lambda)\triangleq\text{Pr}(v | s, h, \lambda)$ for selecting action $v$ in state $s$ conditioned on the history $h\triangleq\{s_i, u_i, v_i\}_{i=1}^{d}$ of $d$ latest interactions where $u_i$ is the action taken by our agent in the $i$-th step. These distributions are parameterized by $\lambda$, which abstracts the actual parametric form of the opponent's behavior; this abstraction provides practitioners the flexibility in choosing the most suitable degree of parameterization. For example, $\lambda$ can simply be a set of multinomial distributions $\lambda\triangleq\{\theta_{sh}^{v}\}$ such that $p_{sh}^{v}(\lambda) \triangleq \theta_{sh}^{v}$ if no prior domain knowledge is available. Otherwise, the domain knowledge can be exploited to produce a fine-grained representation of $\lambda$; at the same time, $\lambda$ can be made compact to generalize the opponent's behavior across different states (e.g., Section~\ref{sect:exp}). 

The opponent's behavior can be learned by monitoring the belief $b(\lambda) \triangleq \text{Pr}(\lambda)$ over all possible $\lambda$. In particular, the belief (or probability density) $b(\lambda)$ is updated at each step based on the history $h \circ \langle s, u, v \rangle$ of $d+1$ latest interactions (with $\langle s, u, v \rangle$ being the most recent one) using Bayes' theorem:
\begin{eqnarray}
b_{sh}^{v}(\lambda) &\propto& p_{sh}^{v}(\lambda)\ b(\lambda)\ . \label{eq:1}
\end{eqnarray}
Let $\bar{s} = (s, h)$ denote an information state that consists of the current state and the history of $d$ latest interactions. When the opponent's behavior is stationary (i.e., $d = 0$), it follows that $\bar{s} = s$. For ease of notations, the main results of our work (in subsequent sections) are presented only for the case where $d = 0$ (i.e., $\bar{s} = s$); extension to the general case just requires replacing $s$ with $\bar{s}$. In this case, \eqref{eq:1} can be re-written as
\begin{eqnarray}
b_{s}^{v}(\lambda) &\propto& p_{s}^{v}(\lambda)\ b(\lambda)\ . \label{eq:2}
\end{eqnarray}
The key difference between our Bayesian modeling paradigm and FDM  \cite{Poupart2006} is that we do not require $b(\lambda)$ and $p_{s}^{v}(\lambda)$ to be, respectively, Dirichlet prior and multinomial likelihood where Dirichlet is a conjugate prior for multinomial. In practice, such a conjugate prior is desirable because the  posterior $b_{s}^{v}$ belongs to the same Dirichlet family as the prior $b$, thus making the belief update tractable and the Bayes-optimal policy efficient to be derived. Despite its computational convenience, this conjugate prior restricts the practitioners from exploiting their domain knowledge to design more informed priors (e.g., see Section~\ref{sect:exp}). Furthermore, this turns out to be an overkill just to make the belief update tractable. In particular, we show in Theorem~\ref{thm:1} below that, without assuming any specific parametric form of the initial prior, the posterior belief can still be tractably represented even though they are not necessarily conjugate distributions. This is indeed sufficient to guarantee and derive a tractable representation of the Bayes-optimal policy using a finite set of parameters, as shall be seen later in Section~\ref{sect:exact}.
\begin{theorem}
\label{thm:1}
If the initial prior $b$ can be represented exactly using a finite set of parameters, then the posterior $b'$ conditioned on a sequence of observations $\{(s_i, v_i)\}_{i=1}^{n'}$ can also be represented exactly in parametric form.
\end{theorem}
\emph{Proof Sketch}. 
From \eqref{eq:2}, we can prove by induction on $n'$
that
\begin{eqnarray}
b'(\lambda) &\propto& \Phi(\lambda) b(\lambda) \label{eq:3}\\
\Phi(\lambda)&\triangleq&\prod_{s\in S}\prod_{v\in V} p_s^v(\lambda)^{\psi_{s}^v}\label{eq:yt}
\end{eqnarray}
where $\psi_{s}^{v} \triangleq\sum_{i=1}^{n'} \delta_{sv}(s_i, v_i)$ and $\delta_{sv}$ is the Kronecker delta function that returns $1$ if $s = s_i$ and $v = v_i$, and $0$ otherwise\footnote{Intuitively, $\Phi(\lambda)$ can be interpreted as the likelihood of observing each pair $(s,v)$ for $\psi_{s}^v$ times while interacting with an opponent whose behavior is parameterized by $\lambda$.}.
From \eqref{eq:3}, it is clear that $b'$ can be represented by a set of parameters $\{\psi_{s}^{v}\}_{s,v}$ and the finite representation of $b$. Thus, belief update is performed simply by incrementing the hyper-parameter $\psi_s^v$ according to each observation $(s, v).\ \ \ \Box$

\section{Interactive Bayesian RL (I-BRL)}
\label{sect:ibrl}
In this section, we first extend the proof techniques used in \cite{Poupart2006} to theoretically derive the agent's Bayes-optimal policy against the general class of parametric models and model priors of the opponent's behavior (Section~\ref{sect:model}). In particular, we show that the derived Bayes-optimal policy can also be represented exactly using a finite number of parameters. Based on our derivation,
a naive algorithm can be devised to compute the exact parametric form of the Bayes-optimal policy (Section~\ref{sect:exact}). Finally, we present a practical algorithm to efficiently approximate this Bayes-optimal policy in polynomial time (with respect to the size of the environment model) (Section~\ref{sect:approx}).    

Formally, an agent is assumed to be interacting with its opponent in a stochastic environment modeled as a tuple $(S, U, V, \{r_s\}, \{p_{s}^{uv}\}, \{p_{s}^{v}(\lambda)\}, \phi)$ where $S$ is a finite set of states, $U$ and $V$ are sets of actions available to the agent and its opponent, respectively. In each stage, the immediate payoff $r_s(u, v)$ to our agent depends on the joint action $(u, v) \in U \times V$ and the current state $s \in S$. The environment then transitions to a new state $s'$ with probability $p_{s}^{uv}(s')\triangleq\text{Pr}(s' | s, u, v)$ and the future payoff (in state $s'$) is discounted by a constant factor $0 < \phi < 1$, and so on. Finally, as described in Section~\ref{sect:model}, the opponent's latent behavior $\{p_{s}^{v}(\lambda)\}$ can be selected from the general class of parametric models and model priors, which subsumes FDM (i.e., independent multinomials with separate Dirichlet priors). 

Now, let us recall that the key idea underlying the notion of Bayes-optimality \cite{Duff2003} is to maintain a belief $b(\lambda)$ that represents the uncertainty surrounding the opponent's behavior $\lambda$ in each stage of interaction. Thus, the action selected by the learner in each stage affects both its expected immediate payoff $\mathbb{E}_{\lambda}\hspace{-1mm}\left[\sum_{v}p_{s}^{v}(\lambda)r_{s}(u,v)|b\right]$ 
and the posterior belief state $b_s^v(\lambda)$, the latter of which influences its future payoff and builds in the information gathering option (i.e., exploration). As such, the Bayes-optimal policy can be obtained by maximizing the expected discounted sum of rewards $V_s(b)$:
\begin{equation}
V_s(b) \triangleq\max_{u} \sum_{v}\left\langle p_{s}^{v},b\right\rangle\hspace{-1mm}\left(r_s(u,v) + \phi\sum_{s'}p_{s}^{uv}(s')V_{s'}(b_s^v)\right) \label{eq:4}
\end{equation}
where $\langle a, b\rangle\triangleq\int_{\lambda} a(\lambda)b(\lambda)\mathrm{d}\lambda$.
The optimal policy for the learner is then defined as a function $\pi^{*}$ that maps the belief $b$ to an action $u$ maximizing its expected utility, which can be derived by solving \eqref{eq:4}. To derive our solution, we first re-state two well-known results concerning the augmented belief-state MDP in single-agent RL \cite{Poupart2006}, which also hold straight-forwardly for our general class of parametric models and model priors. 
\begin{theorem}
\label{thm:2}
The optimal value function $V^k$ for $k$ steps-to-go converges to the optimal value function $V$ for infinite horizon as $k \rightarrow \infty$:
$\|V - V^{k+1}\|_{\infty} \leq \phi\|V - V^k\|_{\infty}\ .$
\end{theorem}
\begin{theorem}
\label{thm:3}
The optimal value function $V_s^k(b)$ for $k$ steps-to-go can be represented by a finite set $\Gamma_s^k$ of $\alpha$-functions:
\begin{equation}
V_s^k(b) = \max_{\alpha_s \in \Gamma_s^k} \left\langle\alpha_s, b\right\rangle .  
\label{eq:7}
\end{equation} 
\end{theorem}
Simply put, these results imply that the optimal value $V_s$ in \eqref{eq:4} can be approximated arbitrarily closely by a finite set $\Gamma_s^k$ of piecewise linear $\alpha$-functions $\alpha_s$, as shown in \eqref{eq:7}. Each $\alpha$-function $\alpha_s$ is associated with an action $u_{\alpha_s}$ yielding an expected utility of $\alpha_s(\lambda)$ if the true behavior of the opponent is $\lambda$ and consequently an overall expected reward $\langle\alpha_s, b\rangle$
%
by assuming that, starting from $(s, b)$, the learner selects action $u_{\alpha_s}$ and continues optimally thereafter. In particular, $\Gamma_s^k$ and $u_{\alpha_s}$ can be derived based on a constructive proof of Theorem~\ref{thm:3}. However, due to limited space, we only state the constructive process below. Interested readers are referred to Appendix A for a detailed proof. Specifically, given $\{\Gamma_s^k\}_s$ such that \eqref{eq:7} holds for $k$, it follows (see Appendix A) that
\begin{eqnarray}
V_s^{k+1}(b) = \max_{u,t}\left\langle\alpha^{ut}_s, b\right\rangle
\label{eq:10}
\end{eqnarray} 
where 
$t\triangleq\left(t_{s'v}\right)_{s' \in S, v \in V}$ with $t_{s'v}\in\left\{1,\ldots,\left|\Gamma_{s'}^{k}\right|\right\}$, and
\begin{equation}
\alpha_s^{ut}(\lambda)\triangleq\sum_v p_s^v(\lambda)\Big(r_s(u,v) + \phi\sum_{s'}\alpha_{s'}^{t_{s'v}}(\lambda)p_s^{uv}(s')\Big) \label{eq:11}
\end{equation}
such that $\alpha_{s'}^{t_{s'v}}$ denotes the $t_{s'v}$-th $\alpha$-function in
$\Gamma_{s'}^{k}$.
Setting $\Gamma_s^{k+1} = \{\alpha_s^{ut}\}_{u,t}$ and $u_{\alpha_s^{ut}} = u$, it follows that \eqref{eq:7} also holds for $k + 1$. As a result, the optimal policy $\pi^{*}(b)$ can be derived directly from these $\alpha$-functions by
$\pi^{\ast}(b)\triangleq u_{\alpha_s^{\ast}}$ where
$\alpha_s^{\ast}=\argmax_{\alpha_s^{ut} \in \Gamma_s^{k+1}}
\left\langle\alpha^{ut}_s, b\right\rangle
.$
Thus, constructing $\Gamma_s^{k+1}$ from the previously constructed sets $\{\Gamma_{s}^k\}_s$ essentially boils down to an exhaustive enumeration of all possible pairs $(u, t)$ and the corresponding application of \eqref{eq:11} to compute $\alpha_s^{ut}$. 
Though \eqref{eq:11} specifies a bottom-up procedure constructing $\Gamma_s^{k+1}$ from the previously constructed sets $\{\Gamma_{s'}^k\}_{s'}$ of $\alpha$-functions, it implicitly requires a convenient parameterization for the $\alpha$-functions that is closed under the application of \eqref{eq:11}. To complete this analytical derivation, we present a final result to demonstrate that each $\alpha$-function is indeed of such parametric form. Note that Theorem~\ref{thm:4} below generalizes a similar result proven in \cite{Poupart2006}, the latter of which shows that, under FDM, each $\alpha$-function can be represented by a linear combination of multivariate monomials. A practical algorithm building on our generalized result in Theorem~\ref{thm:4} is presented in Section~\ref{sect:approx}. 
\begin{theorem}
\label{thm:4}
Let \mbox{\boldmath$\Phi$} denote a family of all functions $\Phi(\lambda)$ \eqref{eq:yt}.
Then, the optimal value $V_{s'}^k$ can be represented by a finite set $\Gamma_{s'}^{k}$ of $\alpha$-functions $\alpha_{s'}^j$ for $j = 1,\ldots, |\Gamma_{s'}^k|$:
\begin{equation}
\alpha_{s'}^j(\lambda)=\sum_{i=1}^m c_i \Phi_i (\lambda) \label{eq:12}
\end{equation}
\noindent where $\Phi_i \in \mbox{\boldmath$\Phi$}$. So, each $\alpha$-function $\alpha_{s'}^j$ can be compactly represented by a finite set of parameters $\{c_i\}_{i=1}^m$\footnote{To ease readability, we abuse the notations $\{c_i, \Phi_i\}_{i=1}^m$ slightly:
Each $\alpha_{s'}^j(\lambda)$ should be specified by a different set $\{c_i, \Phi_i\}_{i=1}^m$.}.
\end{theorem}
\emph{Proof Sketch}. We will prove \eqref{eq:12} by induction on $k$\footnote{\label{k0}When $k=0$, \eqref{eq:12} can be verified by letting $c_i = 0$.}. Supposing \eqref{eq:12} holds for $k$. Setting $j=t_{s'v}$ in \eqref{eq:12} results in
\begin{eqnarray}
\alpha_{s'}^{t_{s'v}}(\lambda) = \sum_{i=1}^m c_i\Phi_i(\lambda) \ ,\label{eq:13}
\end{eqnarray}
which is then plugged into \eqref{eq:11} to yield
\begin{eqnarray}
\alpha_s^{ut}(\lambda) = \sum_{v\in V} c_v\Psi_v(\lambda) + \sum_{s'\in S}\sum_{v\in V}\left(\sum_{i=1}^m c_{s'i}^v\Psi_{s'i}^{v}(\lambda)\right) \label{eq:14}
\end{eqnarray}
where $\Psi_v(\lambda) = p_s^v(\lambda)$, $\Psi_{s'i}^{v}(\lambda) = p_s^v(\lambda)\Phi_i(\lambda)$, and the coefficients $c_v = r_s(u,v)$ and $c_{s'i}^v = \phi p_s^{uv}(s')c_i$. It is easy to see that $\Psi_v \in \mbox{\boldmath$\Phi$}$ and $\Psi_{s'i}^{v} \in \mbox{\boldmath$\Phi$}$. So, \eqref{eq:12} clearly holds for $k+1$. We have shown above that, under the general class of parametric models and model priors (Section~\ref{sect:model}), each $\alpha$-function can be represented by a linear combination of arbitrary parametric functions in \mbox{\boldmath$\Phi$}, which subsume multivariate monomials used in \cite{Poupart2006}$. \ \ \ \Box$

\subsection{An Exact Algorithm}
\label{sect:exact}
Intuitively, Theorems~\ref{thm:3} and \ref{thm:4} provide a simple and constructive method for computing the set of $\alpha$-functions and hence, the optimal policy. In step $k + 1$, the sets $\Gamma_s^{k+1}$ for all $s \in S$ are constructed using \eqref{eq:13} and \eqref{eq:14} from $\Gamma_{s'}^k$ for all $s' \in S$, the latter of which are computed previously in step $k$. When $k = 0$ (i.e., base case), see the proof of Theorem~\ref{thm:4} above (i.e., footnote~\ref{k0}). A sketch of this algorithm is shown below:\vspace{1mm}

\noindent{\bf BACKUP}$\displaystyle(s, k + 1)$\\
1. $\displaystyle\Gamma_{s,u}^{\ast} \leftarrow \left\{g(\lambda) \triangleq\sum_{v} c_v \Psi_v(\lambda)\right\}$\\
2. $\displaystyle\Gamma_{s,u}^{v,s'} \leftarrow \left\{g_j(\lambda)\triangleq\sum_{i=1}^m c_{s'i}^v \Psi_{s'i}^{v}(\lambda)\right\}_{j=1,\ldots, |\Gamma_{s'}^k|}$\\
3. $\displaystyle\Gamma_{s,u} \leftarrow \Gamma_{s,u}^{\ast} \oplus \left(\bigoplus_{v,s'}\Gamma_{s,u}^{v,s'}\right)$\footnote{$A \oplus B = \{a + b | a \in A, b \in B\}$.}\\
4. $\displaystyle\Gamma_{s}^{k+1} \leftarrow \bigcup_{u \in U} \Gamma_{s,u}$\vspace{1mm}

\noindent
In the above algorithm, steps~{\bf 1} and {\bf 2} compute the first and second summation terms on the right-hand side of \eqref{eq:14}, respectively. Then, steps~{\bf 3} and {\bf 4} construct $\Gamma_s^{k+1} = \left\{\alpha_s^{ut}\right\}_{u,t}$ using \eqref{eq:14} over all $t$ and $u$, respectively.
Thus, by iteratively computing $\Gamma_{s}^{k+1} = \text{\bf BACKUP}(s, k+1)$ for a sufficiently large value of $k$, $\Gamma_s^{k+1}$ can be used to approximate $V_s$ arbitrarily closely, as shown in Theorem~\ref{thm:2}. However, this naive algorithm is computationally impractical due to the following issues: (a) {\bf $\alpha$-function explosion} $-$ the number of $\alpha$-functions grows doubly exponentially in the planning horizon length, as derived from \eqref{eq:10} and \eqref{eq:11}: $\left|\Gamma_s^{k+1}\right| = \mathcal{O}\hspace{-1mm}\left(\left[\prod_{s'}\left|\Gamma_{s'}^k\right|\right]^{|V|}|U|\right)$, and (b) {\bf parameter explosion} $-$ the average number of parameters used to represent an $\alpha$-function grows by a factor of $\mathcal{O}\hspace{-1mm}\left(|S||V|\right)$, as manifested in \eqref{eq:14}. The practicality of our approach therefore depends crucially on how these issues are resolved, as described next.
\subsection{A Practical Approximation Algorithm}
\label{sect:approx}
In this section, we introduce practical modifications of the {\bf BACKUP} algorithm by addressing the above-mentioned issues. We first address the issue of {\bf $\alpha$-function explosion} by generalizing discrete POMDP's PBVI solver \cite{Pineau2003} to be used for our augmented belief-state MDP: Only the $\alpha$-functions that yield optimal values for a sampled set of reachable beliefs $B_s = \{b_s^1,b_s^2, \cdots, b_s^{|B_s|}\}$ are computed (see the modifications in steps {\bf 3} and {\bf 4} of the {\bf PB-BACKUP} algorithm). The resulting algorithm is shown below:\vspace{1mm}

\noindent{\bf PB-BACKUP}$\displaystyle(B_s = \{b_s^1, b_s^2, \cdots, b_s^{|B_s|}\}, s, k + 1)$\\
1. $\displaystyle \Gamma_{s,u}^{*} \leftarrow \left\{g(\lambda) \triangleq \sum_{v} c_v \Psi_v(\lambda)\right\}$\\
2. $\displaystyle \Gamma_{s,u}^{v,s'} \leftarrow \left\{g_j(\lambda) \triangleq \sum_{i=1}^m c_{s'i}^v \Psi_{s'i}^{v}(\lambda)\right\}_{j=1,\ldots, |\Gamma_{s'}^k|}$\\
3. $\displaystyle \Gamma_{s,u}^{i} \leftarrow \left\{g + \sum_{s',v}\argmax_{g_j \in \Gamma_{s,u}^{v,s'}}\left\langle g_j,b_s^i\right\rangle\right\}_{g \in \Gamma_{s,u}^{*}}$\\
4. $\displaystyle \Gamma_{s}^{k+1} \leftarrow \left\{g_i \triangleq \argmax_{g \in \Gamma_{s,u}^{i}}\left\langle g,b_s^i\right\rangle\right\}_{i=1,\ldots,|B_s|}$ \vspace{1mm}

\noindent
Secondly, to address the issue of {\bf parameter explosion}, each $\alpha$-function is projected onto a fixed number of basis functions to keep the number of parameters from growing exponentially. This projection is done after each {\bf PB-BACKUP} operation, hence always keeping the number of parameters fixed (i.e., one parameter per basis function). In particular, since each $\alpha$-function is in fact a linear combination of functions in \mbox{\boldmath$\Phi$} (Theorem~\ref{thm:4}), it is natural to choose these basis functions from \mbox{\boldmath$\Phi$}\footnote{See Appendix B for other choices.}. Besides, it is easy to see from \eqref{eq:3} that each sampled belief $b_s^i$ can also be written as
\begin{eqnarray}
b_s^i(\lambda) &=& \eta\Phi_s^i(\lambda)b(\lambda) \label{eq:15}
\end{eqnarray}
where $b$ is the initial prior belief, $\eta = 1/\langle\Phi_s^i, b\rangle$, and $\Phi_s^i \in \mbox{\boldmath$\Phi$}$. For convenience, these $\{\Phi_s^i\}_{i=1,\ldots,|B_s|}$ are selected as basis functions. Specifically, after each {\bf PB-BACKUP} operation, each $\alpha_s \in \Gamma_{s}^k$ is projected onto the function space defined by $\{\Phi_s^i\}_{i=1,\ldots,|B_s|}$. This projection is then cast as an optimization problem that minimizes the squared difference $J(\alpha_s)$ between the $\alpha$-function and its projection with respect to the sampled beliefs in $B_s$:
\begin{eqnarray}
J(\alpha_s) \triangleq \frac{1}{2}\sum_{j=1}^{|B_s|}\left(\left\langle\alpha_s,b_s^j\right\rangle - \sum_{i=1}^{|B_s|}c_i\left\langle\Phi_s^i,b_s^j\right\rangle\right)^{\hspace{-1mm}2} . \label{eq:15c}
\end{eqnarray}
This can be done analytically by letting $\displaystyle\frac{\partial J(\alpha_s)}{\partial c_i} = 0$ and solving for $c_i$, which is equivalent to solving a linear system $Ax = d$ where $x_i = c_i$, $A_{ji} = \sum_{k=1}^{|B_s|}\left\langle\Phi_s^i,b_s^k\right\rangle\left\langle\Phi_s^j,b_s^k\right\rangle$ and $d_j = \sum_{k=1}^{|B_s|}\left\langle\Phi_s^j,b_s^k\right\rangle\left\langle\alpha_s,b_s^k\right\rangle$. Note that this projection works directly with the values $\left\langle\alpha_s,b_s^j\right\rangle$ instead of the exact parametric form of $\alpha_s$ in \eqref{eq:12}. This allows for a more compact implementation of the {\bf PB-BACKUP} algorithm presented above: Instead of maintaining the exact parameters that represent each of the immediate functions $g$, only their evaluations at the sampled beliefs $B_s = \left\{b_s^1, b_s^2, \cdots, b_s^{|B_s|}\right\}$ need to be maintained. In particular, the values of $\left\{\left\langle g,b_s^i\right\rangle\right\}_{i=1,\ldots,|B_s|}$ can be estimated as follows:
\begin{eqnarray}
\left\langle g,b_s^i\right\rangle &=& \eta\int_{\lambda}g(\lambda)\Phi_s^i(\lambda)b(\lambda)\mathrm{d}\lambda \nonumber\\
&\approx& \frac{\sum_{j=1}^n g(\lambda^{j})\Phi_s^i(\lambda^{j})}{\sum_{j=1}^n \Phi_s^i(\lambda^{j})} \label{eq:16}
\end{eqnarray}
where $\left\{\lambda^{j}\right\}_{j=1}^n$ are samples drawn from the initial prior $b$. During the online execution phase, \eqref{eq:16} is also used to compute the expected payoff for the $\alpha$-functions evaluated at the current belief $b'(\lambda) = \eta\Phi(\lambda)b(\lambda)$:
\begin{eqnarray}
\left\langle\alpha_s,b'\right\rangle \approx \frac{\sum_{j=1}^n\Phi(\lambda^{j})\sum_{i=1}^{|B_s|} c_i \Phi_s^i(\lambda^{j})}{\sum_{j=1}^n\Phi(\lambda^{j})}\ . \label{eq:17}
\end{eqnarray}
So, the real-time processing cost of evaluating each $\alpha$-function's expected reward at a particular belief is $\mathcal{O}(|B_s|n)$. Since the sampling of $\{b_s^i\}$, $\{\lambda^{j}\}$ and the computation of $\left\{\sum_{i=1}^{|B_s|}c_i\Phi_s^i(\lambda^{j})\right\}$ can be performed in advance, this $\mathcal{O}(|B_s|n)$ cost is further reduced to $\mathcal{O}(n)$, which makes the action selection incur $\mathcal{O}(|B_s|n)$ cost in total. This is significantly cheaper as compared to the total cost $\mathcal{O}(nk|S|^2|U||V|)$ of online sampling and re-estimating $V_s$ incurred by BPVI \cite{Chalkiadakis2003}. Also, note that since the offline computational costs in steps {\bf 1} to {\bf 4} of {\bf PB-BACKUP$(B_s, s, k+1)$} and the projection cost, which is cast as the cost of solving a system of linear equations, are always polynomial functions of the interested variables (e.g., $|S|, |U|, |V|, n, |B_s|$), the optimal policy can be approximated in polynomial time.

\section{Experiments and Discussion}
\label{sect:exp}
\begin{figure}[h!]
\hspace{15mm}\includegraphics[height=3cm]{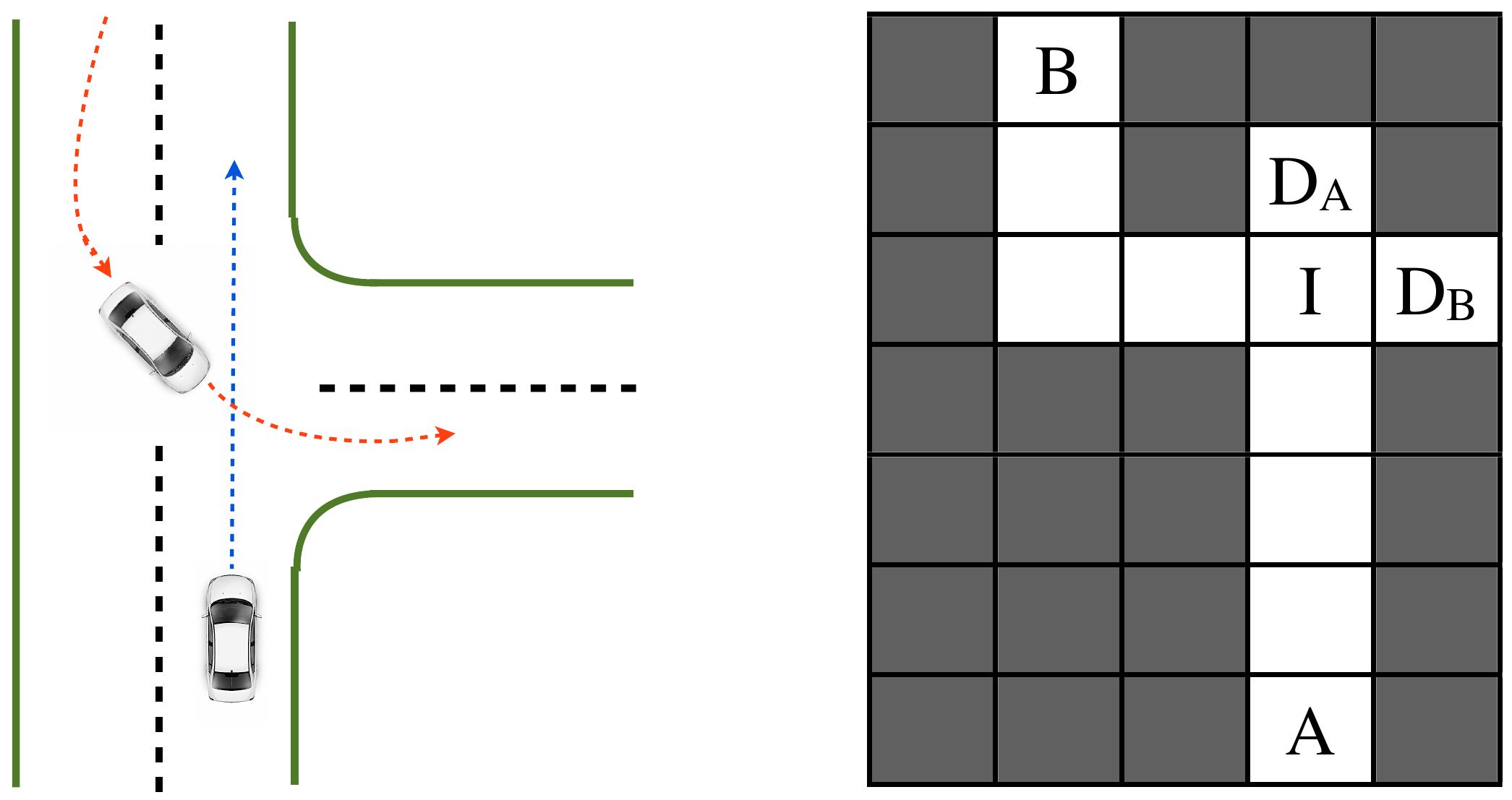}
\caption{(Left) A near-miss accident during the $2007$ DARPA Urban Challenge, and (Right) the discretized environment: $\mathrm{A}$ and $\mathrm{B}$ move towards destinations $\mathrm{D}_{\mathrm{A}}$ and $\mathrm{D}_{\mathrm{B}}$ while avoiding collision at $\mathrm{I}$. Shaded areas are not passable.}
\label{fig:inav}
\end{figure}

In this section, a realistic scenario of intersection navigation is modeled as a stochastic game \cite{Yi12}; it is inspired from a near-miss accident during the $2007$ DARPA Urban Challenge. Considering the traffic situation illustrated in Fig.~\ref{fig:inav} where two autonomous vehicles (marked A and B) are about to enter an intersection (I), the road segments are discretized into a uniform grid with cell size $5$~m $\times\ 5$~m and the speed of each vehicle is also discretized uniformly into $5$ levels ranging from $0$~m/s to $4$~m/s. So, in each stage, the system's state can be characterized as a tuple $\{P_{\mathrm{A}}, P_{\mathrm{B}}, S_{\mathrm{A}}, S_{\mathrm{B}}\}$ specifying the current positions ($P$) and velocities ($S$) of A and B, respectively. In addition, our vehicle (A) can either accelerate ($+1$~m/s$^2$), decelerate ($-1$~m/s$^2$), or maintain its speed ($+0$~m/s$^2$) in each time step while the other vehicle (B) changes its speed based on a parameterized reactive model \cite{Gipps1981}:
\begin{eqnarray}
v_{\text{safe}} &=& S_{\mathrm{B}} + \frac{\text{Distance}(P_{\mathrm{A}}, P_{\mathrm{B}}) - \tau S_{\mathrm{B}}}{S_{\mathrm{B}}/d + \tau} \nonumber \\
v_{\text{des}} &=& \min(4, S_{\mathrm{B}} + a, v_{\text{safe}}) \nonumber\\
S^{'}_{\mathrm{B}} &\sim& \text{Uniform}(\max(0, v_{\text{des}} - \sigma a), v_{\text{des}})\ . \nonumber
\end{eqnarray}
In this model, the driver's acceleration $a \in [0.5~\mbox{m/s$^2$}, 3~\mbox{m/s$^2$}]$, deceleration $d \in [-3~\mbox{m/s$^2$}, -0.5~\mbox{m/s$^2$}]$, reaction time $\tau \in [0.5\mbox{s}, 2\mbox{s}]$, and imperfection $\sigma \in [0, 1]$ are the unknown parameters distributed uniformly within the corresponding ranges. This parameterization can cover a variety of drivers' typical behaviors, as shown in a preliminary study. For a further understanding of these parameters, the readers are referred to \cite{Gipps1981}. Besides, in each time step, each vehicle $\mathrm{X} \in \{\mathrm{A}, \mathrm{B}\}$ moves from its current cell $P_\mathrm{X}$ to the next cell $P^{'}_\mathrm{X}$ with probability $1/t$ and remains in the same cell with probability $1 - 1/t$ where $t$ is the expected time to move forward one cell from the current position with respect to the current speed (e.g., $t = 5/S_\mathrm{X}$). Thus, in general, the underlying stochastic game has $6 \times 6 \times 5 \times 5 = 900$ states (i.e., each vehicle has $6$ possible positions and $5$ levels of speed), which is significantly larger than the settings in previous experiments. In each state, our vehicle has $3$ actions, as mentioned previously, while the other vehicle has $5$ actions corresponding to $5$ levels of speed according to the reactive model.

The goal for our vehicle in this domain is to learn the other vehicle's reactive model and adjust its navigation strategy accordingly such that there is no collision and the time spent to cross the intersection is minimized. To achieve this goal, we penalize our vehicle in each step by $-1$ and reward it with $50$ when it successfully crosses the intersection. If it collides with the other vehicle (at $\mathrm{I}$), we penalize it by $-250$. The discount factor is set as $0.99$. We evaluate the performance of I-BRL in this problem against $100$ different sets of reactive parameters (for the other vehicle) generated uniformly from the above ranges. Against each set of parameters, we run $20$ simulations ($h = 100$ steps each) to estimate our vehicle's average performance\footnote{After our vehicle successfully crosses the intersection, the system's state is reset to the default state in Fig.~\ref{fig:inav} (Right).} $R$. In particular, we compare our algorithm's average performance against the average performance of a fully informed vehicle ({\bf Upper Bound}) who knows exactly the reactive parameters before each simulation, a rational vehicle ({\bf Exploit}) who estimates the reactive parameters by taking the means of the above ranges, and a vehicle employing BPVI \cite{Chalkiadakis2003} ({\bf BPVI}).
\begin{figure}
\includegraphics[height=3.32cm]{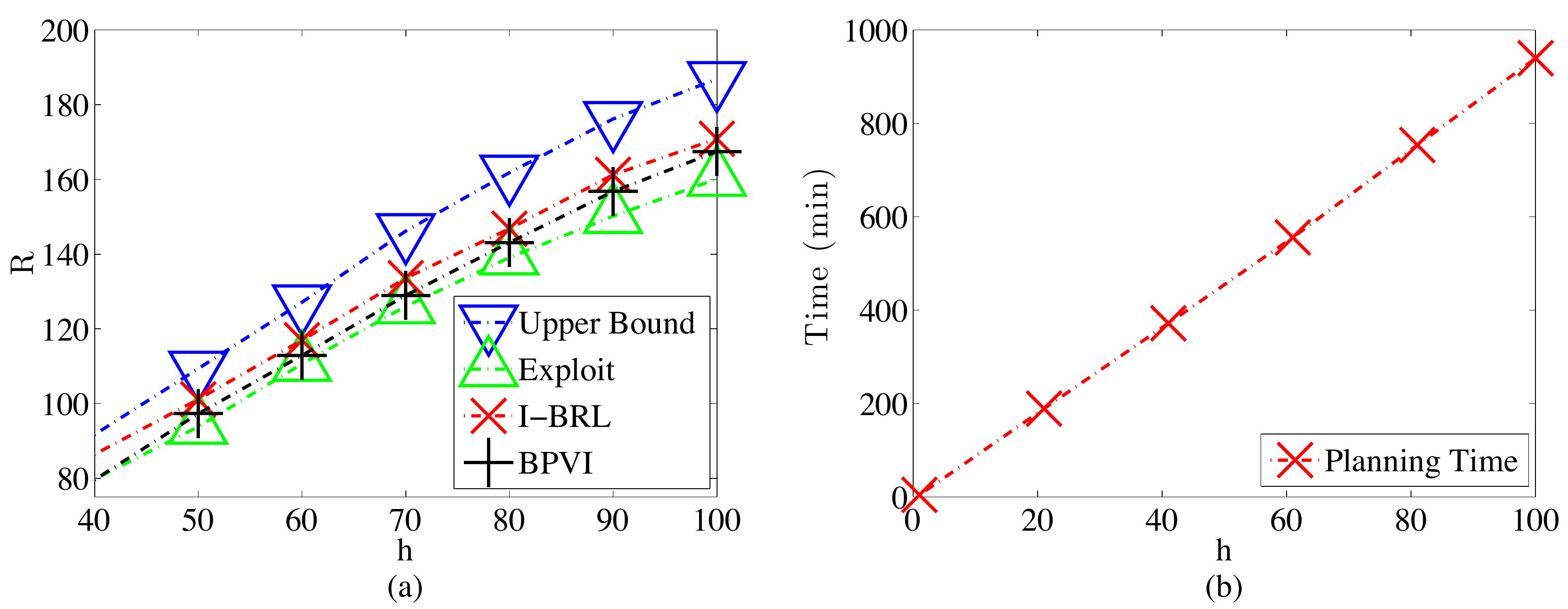}
\caption{(a) Performance comparison between our vehicle (I-BRL), the fully informed, the rational and the BPVI vehicles ($\phi = 0.99$); (b) Our approach's offline planning time.}
\label{fig:inav_result}
\end{figure}

The results are shown in Fig.~\ref{fig:inav_result}a: It can be observed that our vehicle always performs significantly better than both the rational and BPVI-based vehicles. In particular, our vehicle manages to reduce the performance gap between the fully informed and rational vehicles roughly by half. The difference in performance between our vehicle and the fully informed vehicle is expected as the fully informed vehicle always takes the optimal step from the beginning (since it knows the reactive parameters in advance) while our vehicle has to take cautious steps (by maintaining a slow speed) before it feels confident with the information collected during interaction. Intuitively, the performance gap is mainly caused during this initial period of ``caution''. Also, since the uniform prior over the reactive parameters $\lambda = \{a, d, \tau, \sigma\}$ is not a conjugate prior for the other vehicle's behavior model $\theta_s(v) = p(v|s,\lambda)$, the BPVI-based vehicle has to directly maintain and update its belief using FDM: $\lambda = \{\theta_s\}_s$ with $\theta_s = \{\theta_s^v\}_v \sim \text{Dir}(\{n_s^v\}_v)$ (Section~\ref{sect:model}), instead of $\lambda = \{a, d, \tau, \sigma\}$. However, FDM implicitly assumes that $\{\theta_s\}_s$ are statistically independent, which is not true in this case since all $\theta_s$ are actually related by $\{a, d, \tau, \sigma\}$. Unfortunately, BPVI cannot exploit this information to generalize the other vehicle's behavior across different states due to its restrictive FDM (i.e., independent multinomial likelihoods with separate Dirichlet priors), thus resulting in an inferior performance.
\section{Related Works}
\label{sect:related}
In self-interested (or non-cooperative) MARL, there has been several groups of proponents advocating different learning goals, the following of which have garnered substantial support: (a) {\bf Stability} $-$ in self-play or against a certain class of learning opponents, the learners' behaviors converge to an equilibrium; (b) {\bf optimality} $-$ a learner's behavior necessarily converges to the best policy against a certain class of learning opponents; and (c) {\bf security} $-$ a learner's average payoff must exceed the maximin value of the game. For example, the works of \citeauthor{Littman2001}~\shortcite{Littman2001}, \citeauthor{Bianchi2007}~\shortcite{Bianchi2007}, and \citeauthor{Akchurina2009}~\shortcite{Akchurina2009} have focused on (evolutionary) game-theoretic approaches that satisfy the {\bf stability} criterion in self-play. The works of 
\citeauthor{Bowling2001}~\shortcite{Bowling2001}, \citeauthor{Suematsu2002}~\shortcite{Suematsu2002}, and \citeauthor{Tesauro2003}~\shortcite{Tesauro2003} have developed algorithms that address both the {\bf optimality} and {\bf stability} criteria: A learner essentially converges to the best response if the opponents' policies are stationary; otherwise, it converges in self-play. Notably, the work of \citeauthor{Powers2005}~\shortcite{Powers2005} has proposed an approach that provably converges to an $\epsilon$-best response (i.e., {\bf optimality}) against a class of adaptive, bounded-memory opponents while simultaneously guaranteeing a minimum average payoff (i.e., {\bf security}) in single-state, repeated games. 

In contrast to the above-mentioned works that focus on convergence, I-BRL directly optimizes a learner's performance during its course of interaction, which may terminate before it can successfully learn its opponent's behavior. So, our main concern is how well the learner can perform before its behavior converges. 
From a practical perspective, this seems to be a more appropriate goal: In reality, the agents may only interact for a limited period, which is not enough to guarantee convergence, thus undermining the {\bf stability} and {\bf optimality} criteria.
In such a context, the existing approaches appear to be at a disadvantage: (a) Algorithms that focus on {\bf stability} and {\bf optimality} tend to select exploratory actions with drastic effect without considering their huge costs (i.e., poor rewards)  \cite{Chalkiadakis2003}; and (b) though the notion of {\bf security} aims to prevent a learner from selecting such radical actions, the proposed security values (e.g., maximin value) may not always turn out to be tight lower bounds for the optimal performance \cite{Anonymous2012}. Interested readers are referred to \cite{Chalkiadakis2003} and Appendix C for a detailed discussion and additional experiments to compare performances of I-BRL and these approaches, respectively. 

Note that while solving for the Bayes-optimal policy efficiently has not been addressed explicitly in general prior to this paper, we can actually avoid this problem by allowing the agent to act sub-optimally in a bounded number of steps. In particular, the works of \citeauthor{Asmuth2011}~\shortcite{Asmuth2011} and \citeauthor{Mauricio2012}~\shortcite{Mauricio2012} guarantee that, in the worst case, the agent will act nearly approximately Bayes-optimal in all but a polynomially bounded number of steps with high probability. It is thus necessary to point out the difference between I-BRL and these worst-case approaches: We are interested in maximizing the average-case performance with certainty rather than the worst-case performance with some ``high probability" guarantee. Comparing their performances is beyond the scope of this paper.
\section{Conclusion}
\label{sect:ccl}
This paper describes a novel generalization of BRL, called I-BRL, to integrate the general class of parametric models and model priors of the opponent's behavior. As a result, I-BRL relaxes the restrictive assumption of FDM that is often imposed in existing works, thus offering practitioners greater flexibility in encoding their prior domain knowledge of the opponent's behavior. Empirical evaluation shows that I-BRL outperforms a Bayesian MARL approach utilizing FDM called BPVI. I-BRL also outperforms existing MARL approaches focusing on convergence (Section~\ref{sect:related}), as shown in the additional experiments in \cite{Anonymous2012}. To this end, we have successfully bridged the gap in applying BRL to self-interested multi-agent settings.\vspace{1mm}

\noindent
{\bf Acknowledgments.} 
This work was supported by Singapore-MIT Alliance Research \& Technology Subaward Agreements No. $28$ R-$252$-$000$-$502$-$592$ \& No. $33$ R-$252$-$000$-$509$-$592$.

\bibliographystyle{named}
\bibliography{ijcai13}

\ifthenelse{\value{sol}=1}{
\clearpage

\appendix

\section{Proof Sketches for Theorems~\ref{thm:2} and~\ref{thm:3}}
This section provides more detailed proof sketches for Theorems~\ref{thm:2} and~\ref{thm:3} as mentioned in Section~\ref{sect:ibrl}.\vspace{2mm}

{\bf \noindent Theorem 2.} The optimal value function $V^k$ for $k$ steps-to-go converges to the optimal value function $V$ for infinite horizon as $k \rightarrow \infty$:
$
\|V - V^{k+1}\|_{\infty} \leq \phi\|V - V^k\|_{\infty}$.
\vspace{1.5mm}

\noindent\emph{Proof Sketch}. Define $L_s^k(b) = |V_s(b) - V_s^k(b)|$. Using $|\max_a f(a) - \max_a g(a)| \leq \max_a |f(a) - g(a)|$, 
\begin{eqnarray}
L_s^{k+1}(b) &\leq& \phi\max_{u}\sum_{v,s'}\left\langle p_s^v,b\right\rangle p_s^{uv}(s')L_{s'}^{k}(b_s^v) \nonumber\\
&\leq& \phi\max_{u}\sum_{v,s'}\left\langle p_s^v,b\right\rangle p_s^{uv}(s')\|V - V^k\|_{\infty}\nonumber\\
&=& \phi\|V - V^k\|_{\infty} \label{eq:6a}
\end{eqnarray}
Since the last inequality \eqref{eq:6a} holds for every pair $(s,b)$, it follows that $\|V - V^{k+1}\|_{\infty} \leq \phi \|V - V^k\|_{\infty}$.
\vspace{3mm}

{\bf \noindent Theorem 3.} The optimal value function $V_s^k(b)$ for $k$ steps-to-go can be represented as a finite set $\Gamma_s^k$ of $\alpha$-functions:
\begin{equation}
V_s^k(b) = \max_{\alpha_s \in \Gamma_s^k} \left\langle\alpha_s,b\right\rangle \ .\label{eq:7a}
\end{equation} 
\vspace{1.5mm}

\noindent\emph{Proof Sketch}. We give a constructive proof to \eqref{eq:7a} by induction, which shows how $\Gamma_s^k$ can be built recursively. Assuming that \eqref{eq:7a} holds for $k$ \footnote{When $k = 0$, \eqref{eq:7a} can be verified by letting $\alpha_s(\lambda) = 0$}, it can be proven that \eqref{eq:7a} also holds for $k+1$. In particular, it follows from our inductive assumption that the term $V_{s'}^k(b_s^v)$ in \eqref{eq:4} can be rewritten as:
\begin{eqnarray}
V_{s'}^k(b_s^v) &=& \max_{j=1}^{|\Gamma_{s'}^{k}|} \int_{\lambda}\alpha_{s'}^j(\lambda)b_s^v(\lambda)\mathrm{d}\lambda \nonumber\\
&=& \max_{j=1}^{|\Gamma_{s'}^{k}|} \int_{\lambda}\alpha_{s'}^j(\lambda)\frac{p_s^v(\lambda)b(\lambda)}{\left\langle p_s^v,b\right\rangle}\mathrm{d}\lambda \nonumber \\
&=& \left\langle p_s^v,b\right\rangle^{-1} \max_{j=1}^{|\Gamma_{s'}^{k}|} \int_{\lambda}b(\lambda)\alpha_{s'}^j(\lambda)p_s^v(\lambda)\mathrm{d}\lambda \nonumber
\end{eqnarray}
By plugging the above equation into \eqref{eq:4} and using the fact that $r_b^s(u) = \sum_v \left\langle p_s^v,b\right\rangle r_s(u,v)$, 
\begin{equation}
V_s^{k+1}(b) = \max_u \left(r_b^s(u) + \phi\sum_{s',v}\max_{j=1}^{|\Gamma_{s'}^{k}|}p_s^{uv}(s')Q_{s}^{v}(\alpha_{s'}^j, b) \right) \label{eq:9a}
\end{equation}
where $Q_{s}^{v}(\alpha_{s'}^j, b) = \int_{\lambda}\alpha_{s'}^j(\lambda)p_s^v(\lambda)b(\lambda)\mathrm{d}\lambda$. Now, applying the fact that
\begin{eqnarray}
\sum_{s'}\sum_{v} \max_{t_{s'v} = 1}^{|\Gamma_{s'}^k|} A_{s'v}[t_{s'v}] = \max_{t}\sum_{s'}\sum_{v} A_{s'v}[t_{s'v}] \nonumber
\end{eqnarray}
where $A_{s'v}[t_{s'v}] = p_s^{uv}(s')Q_{s}^{v}(\alpha_{s'}^{t_{s'v}}, b)$ and using $r_b^s(u) = \int_{\lambda}b(\lambda)\sum_v p_s^v(\lambda)r_s(u,v)\mathrm{d}\lambda$, \eqref{eq:9a} can be rewritten as
\begin{eqnarray}
V_s^{k+1}(b) = \max_{u,t}\int_{\lambda}b(\lambda)\alpha_s^{ut}(\lambda)\mathrm{d}\lambda \label{eq:10a}
\end{eqnarray}
with $t = \left(t_{s'v}\right)_{s' \in S, v \in V}$ and 
\begin{equation}
\alpha_s^{ut}(\lambda) = \sum_v p_s^v(\lambda)\left(r_s(u,v) + \phi\sum_{s'}\alpha_{s'}^{t_{s'v}}(\lambda)p_s^{uv}(s')\right). \label{eq:11a}
\end{equation}
By setting $\Gamma_s^{k+1} = \{\alpha_s^{ut}\}_{u,t}$ and $u_{\alpha_s^{ut}} = u$, it can be verified that \eqref{eq:7a} also holds for $k + 1$.

\section{Alternative Choice of Basis Functions}

This section demonstrates another theoretical advantage of our framework: The flexibility to customize the general point-based algorithm presented in Section~\ref{sect:approx} into more manageable forms (e.g., simple, easy to implement, etc.) with respect to different choices of basis functions. Interestingly, these customizations often allow the practitioners to trade off effectively between the performance and sophistication of the implemented algorithm: A simple choice of basis functions may (though not necessarily) reduce its performance but, in exchange, bestows upon it a customization that is more computationally efficient and easier to implement. This is especially useful in practical situations where finding a good enough solution quickly is more important than looking for better yet time-consuming solutions.

As an example, we present such an alternative of the basis functions in the rest of this section. In particular, let $\{\lambda^{i}\}_{i=1}^n$ be a set of the opponent's models sampled from the initial belief $b$. Also, let $\Psi_{i}(\lambda)$ denote a function that returns $1$ if $\lambda = \lambda^{i}$, and $0$ otherwise. According to Section~\ref{sect:approx}, to keep the number of parameters from growing exponentially, we project each $\alpha$-function onto $\{\Psi_{i}(\lambda)\}_{i=1}^n$ by minimizing \eqref{eq:15c} or alternatively, the unconstrained squared difference between the $\alpha$-function and its projection:
\begin{eqnarray}
J(\alpha_s) &=& \frac{1}{2}\int_{\lambda}\left(\alpha_s(\lambda) - \sum_{i = 1}^{|B_s|}c_i\Phi_s^i(\lambda)\right)^2\mathrm{d}\lambda\ . \label{eq:15b}
\end{eqnarray}
Now, let us consider \eqref{eq:11}, which specifies the exact solution for \eqref{eq:4} in Section~\ref{sect:ibrl}. Assume that $\alpha_{s'}^{t_{s'v}}(\lambda)$ is projected onto $\{\Psi_{i}(\lambda)\}_{i=1}^n$ by minimizing \eqref{eq:15b}:
\begin{equation}
\bar{\alpha}_{s'}^{t_{s'v}}(\lambda) = \sum_{i=1}^n \Psi_i(\lambda) \varphi_{s'}^{t_{s'v}}(i) \label{eq:18}
\end{equation}
where $\{\varphi_{s'}^{t_{s'v}}(i)\}_i$ are the projection coefficients. According to the general point-based algorithm in Section~\ref{sect:approx}, $\alpha_{s}^{ut}(\lambda)$ is first computed by replacing $\alpha_{s'}^{t_{s'v}}(\lambda)$ with $\bar{\alpha}_{s'}^{t_{s'v}}(\lambda)$ in \eqref{eq:11}:
\begin{equation}
\alpha_s^{ut}(\lambda) = \sum_v p_s^v(\lambda)\left(r_s(u,v) + \phi\sum_{s'}\bar{\alpha}_{s'}^{t_{s'v}}(\lambda)p_s^{uv}(s')\right). \label{eq:19}
\end{equation}
\noindent Then, following \eqref{eq:15b}, $\alpha_s^{ut}(\lambda)$ is projected onto $\{\Psi_i(\lambda)\}_i$ by solving for $\{\varphi_{s}^{ut}(i)\}_i$ that minimize
\begin{eqnarray}
J(\alpha_{s}^{ut}) = \frac{1}{2}\int_{\lambda}\left(\alpha_{s}^{ut}(\lambda) - \sum_{i = 1}^{n}\Psi_i(\lambda)\varphi_{s}^{ut}(i)\right)^2\mathrm{d}\lambda\ . \label{eq:20}
\end{eqnarray} 
The back-up operation is therefore cast as finding $\{\varphi_s^{ut}(i)\}_i$ that minimize \eqref{eq:20}. To do this, define $$L(\lambda) = \frac{1}{2}\left(\alpha_{s}^{ut}(\lambda) - \sum_{i = 1}^{n}\Psi_i(\lambda)\varphi_{s}^{ut}(i)\right)^2$$ and take the corresponding partial derivatives of $L(\lambda)$ with respect to $\{\varphi_s^{ut}(j)\}_j$:
\begin{eqnarray}
\frac{\partial L(\lambda)}{\partial \varphi_{s}^{ut}(j)} = -\left(\alpha_{s}^{ut}(\lambda) - \sum_{i=1}^n\Psi_i(\lambda)\varphi_{s}^{ut}(i)\right)\Psi_j(\lambda)\ . \label{eq:21}
\end{eqnarray}
\noindent From the definition of $\Psi_j(\lambda)$, it is clear that when $\lambda \neq \lambda^{j}$, $\displaystyle\frac{\partial L(\lambda)}{\partial \varphi_{s}^{ut}(j)} = 0$. Otherwise, this only happens when\vspace{-3mm}
\begin{eqnarray}
\alpha_{s}^{ut}(\lambda^{j}) &=& \sum_{i = 1}^n \Psi_i(\lambda^{j})\varphi_s^{ut}(i) \nonumber\\
&=& \varphi_s^{ut}(j) \text{ (by def. of $\Psi_i(\lambda)$)}\label{eq:22}
\end{eqnarray}
\noindent On the other hand, by plugging \eqref{eq:18} into \eqref{eq:19} and using $\Psi_i(\lambda) = 0 \text{ }\forall \lambda \ne \lambda^{i}$, $\alpha_{s}^{ut}(\lambda^{j})$ can be expressed as
\begin{equation}
\alpha_{s}^{ut}(\lambda^{j}) = \sum_{v}p_s^v(\lambda^{j})\left(r_s(u,v) + \phi\sum_{s'}p_s^{uv}(s') \varphi_{s'}^{t_{s'v}}(j)\right)\nonumber
\end{equation}
So, to guarantee that $\displaystyle\frac{\partial L(\lambda)}{\partial \varphi_{s}^{ut}(j)} = 0 \text{ } \forall \lambda, j$ (i.e., minimizing \eqref{eq:20} with respect to $\{\varphi_s^{ut}(j)\}_j$), the values of $\{\varphi_s^{ut}(j)\}_j$ can be set as (from \eqref{eq:22} and the above equation)
\begin{eqnarray}
\varphi_s^{ut}(j) = \sum_{v}p_s^v(\lambda^{j})\left(r_s(u,v) + \phi\sum_{s'}p_s^{uv}(s') \varphi_{s'}^{t_{s'v}}(j)\right). \nonumber 
\end{eqnarray}
\noindent Surprisingly, this equation specifies exactly the $\alpha$-vector back-up operation for the discrete version of \eqref{eq:4}:
\begin{equation}
V_s(\hat{b}) = \max_{u} \sum_{v}\left\langle p_{s}^{v},\hat{b}\right\rangle\left(r_s(u,v) + \phi\sum_{s'}p_{s}^{uv}(s')V_{s'}(\hat{b}_s^v)\right) \label{eq:23}
\end{equation}
where $\hat{b}$ is the discrete distribution over the set of samples $\{\lambda^{j}\}_j$ (i.e., $\sum_j\hat{b}(\lambda^{j}) = 1$). This implies that by choosing $\{\Psi_i(\lambda)\}_i$ as our basis functions, finding the corresponding ``projected'' solution for \eqref{eq:4} is identical to solving \eqref{eq:23}, which can be easily implemented using any of the existing discrete POMDP solvers (e.g., \cite{Pineau2003}). 

\section{Additional Experiments}

This section provides additional evaluations of our proposed I-BRL framework, in comparison with existing works in MARL, through a series of stochastic games which are adapted and simplified from the testing benchmarks used in \cite{Yi12,Poupart2006}. In particular, I-BRL is evaluated in two small yet typical application domains that are widely used in the existing works (Sections~\ref{sect:chain}, \ref{sect:ipd}).

\subsection{Multi-Agent Chain World}
\label{sect:chain}
In this problem, the system consists of a chain of $5$ states and $2$ agents; each agent has $2$ actions $\{a, b\}$. In each stage of interaction, both agents will move one step forward or go back to the initial state depending on whether they coordinate on action $a$ or $b$, respectively \cite{Poupart2006}. In particular, the agents will receive an immediate reward of $10$ for coordinating on $a$ in the last state and $2$ for coordinating on $b$ in any state except the first one. Otherwise, the agents will remain in the current state and get no reward (Fig.~\ref{fig:chain_world}). After each step, their payoffs are discounted by a constant factor of $0 < \phi < 1$.

\begin{figure}[h!]
\vspace{-4mm}
\hspace{5mm}\includegraphics[height=3.6cm,width=7cm]{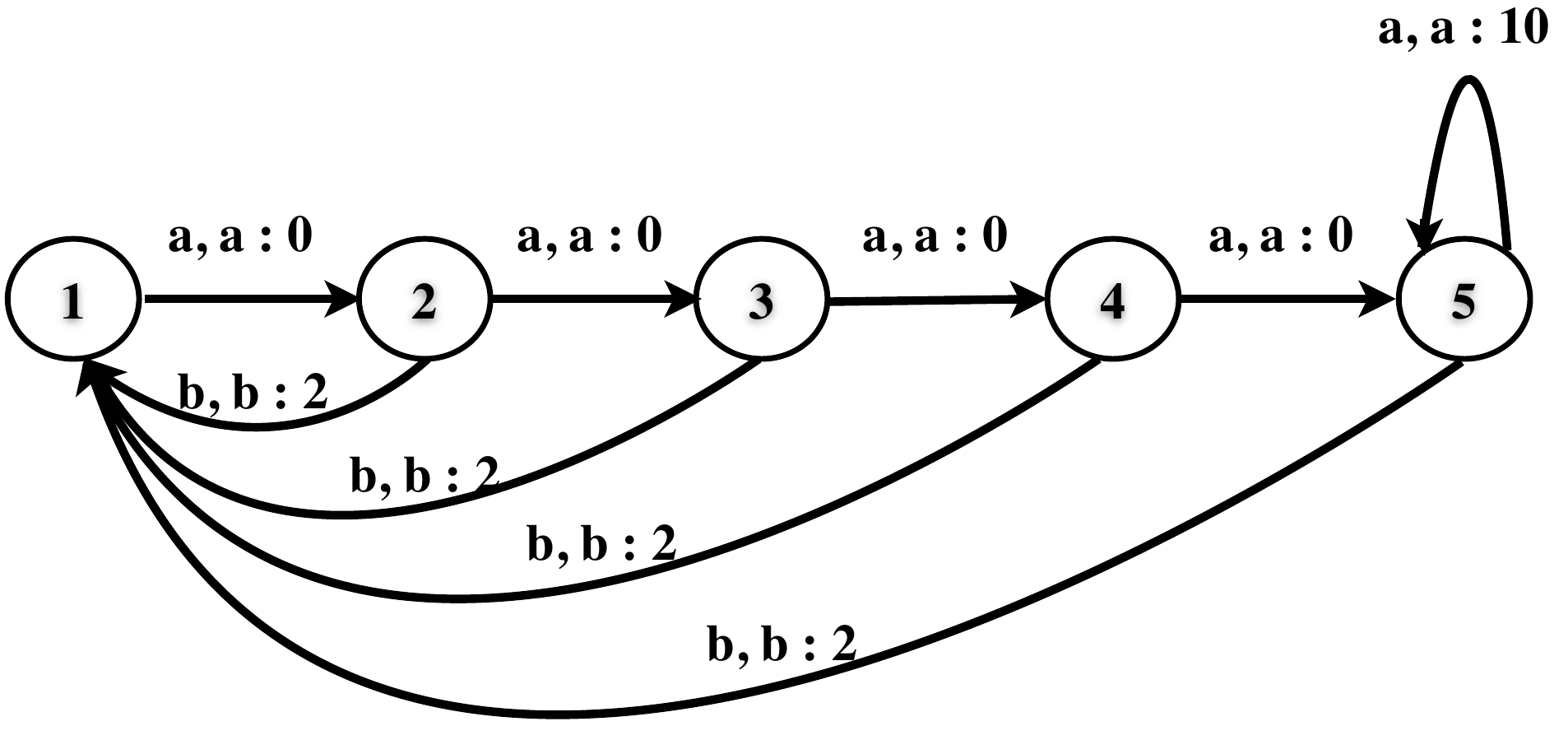}\vspace{-4mm}
\caption{Multi-Agent Chain World Problem.}\vspace{-2mm}
\label{fig:chain_world}
\end{figure}
In this experiment, we compare the performance of I-BRL with the state-of-the-art frameworks in MARL which include BPVI \cite{Chalkiadakis2003}, Hyper-Q \cite{Tesauro2003}, and Meta-Strategy \cite{Powers2005}. Among these works, BPVI is the most relevant to I-BRL (Section~\ref{sect:intro}). Hyper-Q simply extends Q-learning into the context of multi-agent learning. Meta-Strategy, by default, plays the best response to the empirical estimation of the opponent's behavior and occasionally switches to the maximin strategy when its accumulated reward falls below the maximin value of the game. In particular, we compare the average performance of these frameworks when tested against $100$ different opponents whose behaviors are modeled as a set of probabilities $\theta_s = \{\theta_s^v\}_v \sim \mathrm{Dir}(\{n_s^v\}_v)$ (i.e., of selecting action $v$ in state $s$). These opponents are independently and randomly generated from these Dirichlet distributions with the parameters $n_s^v = 1/|V|$. Then, against each opponent, we run $20$ simulations ($h = 100$ steps each) to evaluate the performance $R$ of each framework. The results show that I-BRL significantly outperforms the others (Fig.~\ref{fig:chain_world_result}a).

\begin{figure}[h!]
\vspace{1mm}
\begin{tabular}{cc}
\hspace{-5.5mm}\includegraphics[height=3.65cm,width=4.6cm]{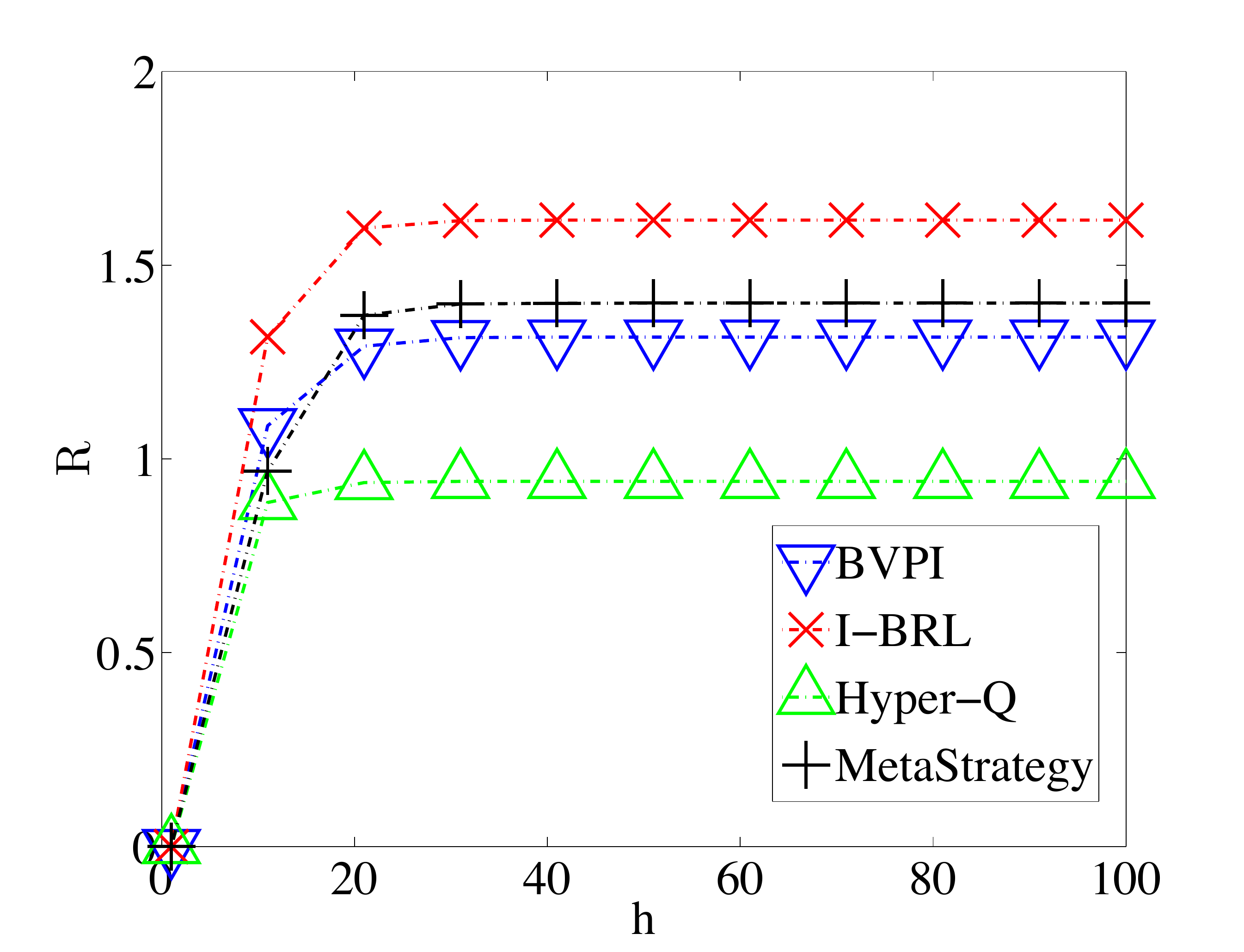} & \hspace{-5.5mm}\includegraphics[height=3.65cm,width=4.6cm]{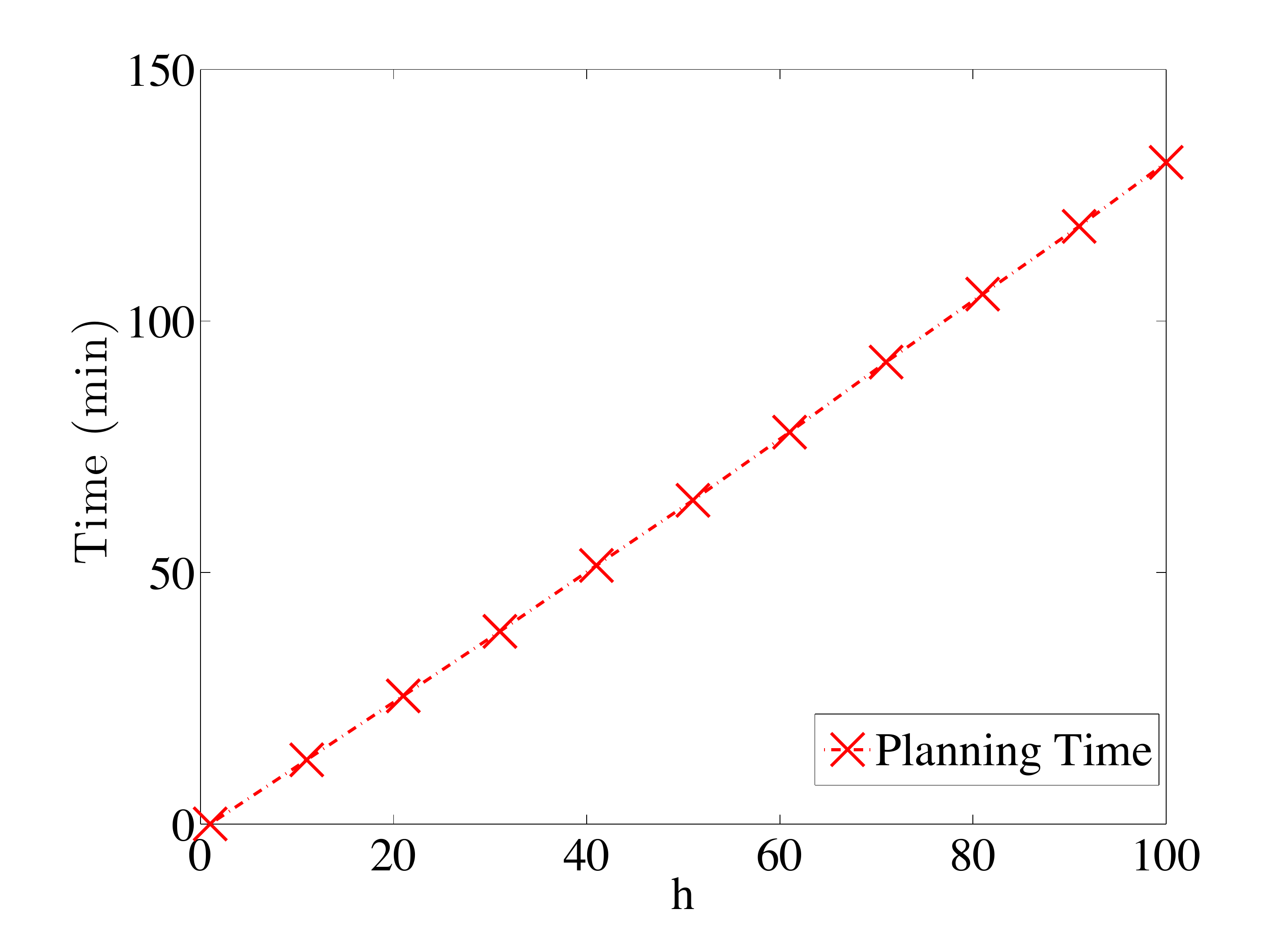}\vspace{-1mm}\\
\hspace{-5.5mm}{\footnotesize (a)} & \hspace{-5.5mm}{\footnotesize (b)}\vspace{-3mm}
\end{tabular}
\caption{(a) Performance comparison between I-BRL, BPVI, Hyper-Q, and Meta-Strategy ($\phi = 0.75$), and (b) I-BRL's offline planning time.}\vspace{-2mm}
\label{fig:chain_world_result}
\end{figure}
From these results, BPVI's inferior performance, as compared to I-BRL, is expected because BPVI, as mentioned in Section~\ref{sect:intro}, relies on a sub-optimal myopic information-gain function \cite{Dearden1998}, thus underestimating the risk of moving forward and forfeiting the opportunity to go backward to get more information and earn the small reward. Hence, in many cases, the chance of getting big reward (before it is severely discounted) is accidentally over-estimated due to BPVI's lack of information. As a result, this makes the expected gain of moving forward insufficient to compensate for the risk of doing so. Besides, it is also expected that Hyper-Q's and Meta-Strategy's performance are worse than I-BRL's since they primarily focus on the criteria of {\bf optimality} and {\bf security}, which put them at a disadvantage in the context of our work, as explained previously in Section~\ref{sect:intro}. Notably, in this case, the maximin value of Meta-Strategy in the first state is vacuously equal to $0$, which is effectively a lower bound for any algorithms. In contrast, I-BRL directly optimizes the agent's expected utility by taking into account its current belief and all possible sequences of future beliefs (see \eqref{eq:4}). As a result, our agent behaves cautiously and always takes the backward action until it has sufficient information to guarantee that the expected gain of moving forward is worth the risk of doing so. In addition, I-BRL's online processing cost is also significantly less expensive than BPVI's: I-BRL requires only $2.5$ hours to complete $20$ simulations ($100$ steps each) against $100$ opponents while BPVI requires $4.2$ hours. In exchange for this speed-up, I-BRL spends a few hours of offline planning (Fig.~\ref{fig:chain_world_result}b), which is a reasonable trade-off considering how critical it is for an agent to meet the real-time constraint during interaction.
\subsection{Iterated Prisoner Dilemma (IPD)}
\label{sect:ipd}
The IPD is an iterative version of the well-known one-shot, two-player game known as the Prisoner Dilemma \cite{Yi12}, in which each player attempts to maximize its reward by cooperating (C) or betraying (B) the other. Unlike the one-shot game, the game in IPD is played repeatedly and each player knows the history of his opponent's moves, thus having the opportunity to predict the opponent's behavior based on past interactions. In each stage of interaction, the agents will get a reward of $3$ or $1$ depending on whether they mutually cooperate or betray each other, respectively. In addition, the agent will get no reward if it cooperates while his opponent betrays; conversely, it gets a reward of $5$ for betraying while the opponent cooperates.

\begin{figure}[h!]
\vspace{-3mm}
\begin{tabular}{cc}
\hspace{-5.5mm}\includegraphics[height=3.65cm,width=4.6cm]{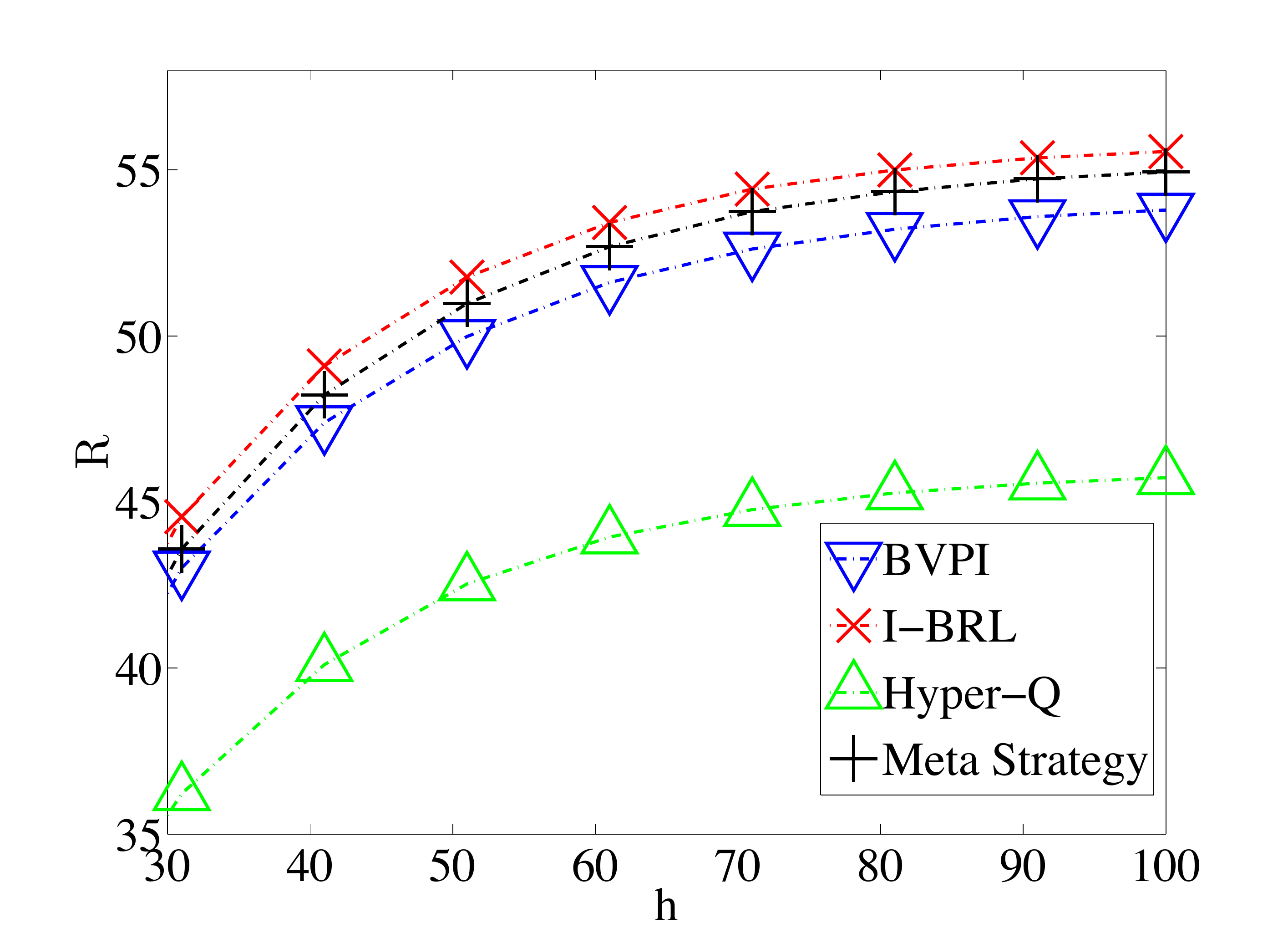} & \hspace{-5.5mm}\includegraphics[height=3.65cm,width=4.6cm]{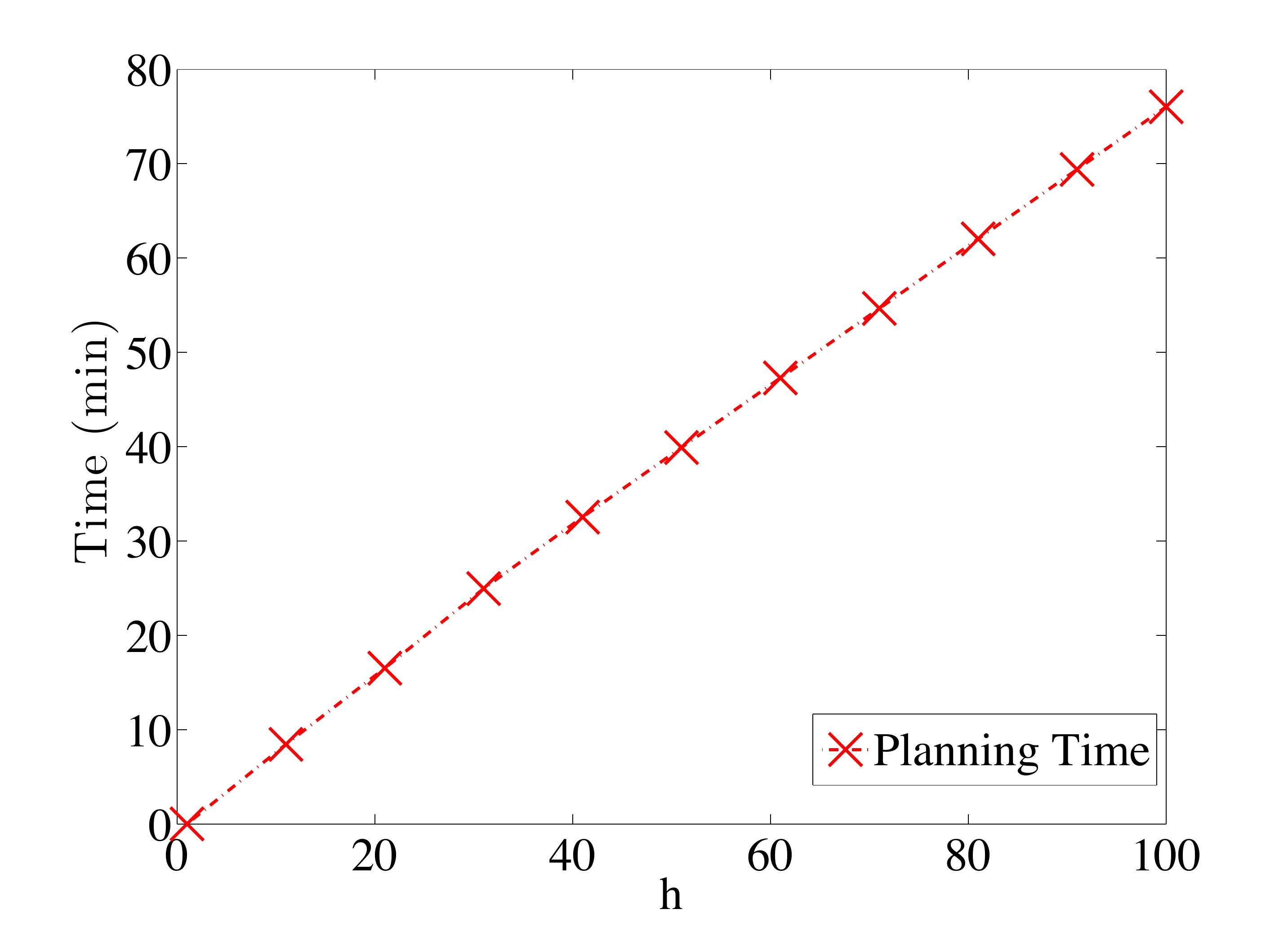}\vspace{-1mm}\\
\hspace{-5.5mm}{\footnotesize (a)} & \hspace{-5.5mm}{\footnotesize (b)}\vspace{-3mm}
\end{tabular}
\caption{(a) Performance comparison between I-BRL, BPVI, Hyper-Q and Meta-Strategy ($\phi = 0.95$); (b) I-BRL's offline planning time.}\vspace{-2mm}
\label{fig:ipd_result}
\end{figure}

In this experiment, the opponent is assumed to make his decision based on the last step of interaction (e.g., adaptive, memory-bounded opponent). Thus, its behavior can be modeled as a set of conditional probabilities $\theta_{\bar{s}}= \{\theta_{\bar{s}}^v\}_v \sim \mathrm{Dir}(\{n_{\bar{s}}^v\}_v)$ where $\bar{s} = \{a_{-1}, o_{-1}\}$ encodes the agent's and its opponent's actions $a_{-1}, o_{-1} \in \{\mathrm{B}, \mathrm{C}\}$ in the previous step. Similar to the previous experiment (Section~\ref{sect:chain}), we compare the average performance of I-BRL, BPVI \cite{Chalkiadakis2003}, Hyper-Q \cite{Tesauro2003}, and Meta-Strategy \cite{Powers2005} when tested against $100$ different opponents randomly generated from the Dirichlet priors. The results are shown in Figs.~\ref{fig:ipd_result}a and \ref{fig:ipd_result}b. From these results, it can be clearly observed that I-BRL also outperforms BPVI and the other methods in this experiment, which is consistent with our explanations in the previous experiment. Also, in terms of the online processing cost, I-BRL only requires $1.74$ hours to complete all the simulations while BPVI requires $4$ hours.
}{}

\end{document}